\documentclass[11pt]{article}

\usepackage[letterpaper,margin=1in]{geometry}
\usepackage[T1]{fontenc}
\usepackage{lmodern}
\usepackage{microtype}
\usepackage{amsmath}
\usepackage{amssymb}
\usepackage{amsthm}
\usepackage{booktabs}
\usepackage{graphicx}
\usepackage[round]{natbib}
\usepackage{xcolor}
\usepackage[colorlinks=true,linkcolor=blue!60!black,citecolor=blue!60!black,
  urlcolor=blue!60!black,hypertexnames=false]{hyperref}

\theoremstyle{plain}
\newtheorem{theorem}{Theorem}
\newtheorem{lemma}{Lemma}
\newtheorem{corollary}{Corollary}

\theoremstyle{definition}

\theoremstyle{remark}
\newtheorem*{remark}{Remark}

\theoremstyle{plain}
\makeatletter
\newenvironment{thmn}[2][]{%
  \renewcommand{\thetheorem}{#2}%
  \if\relax\detokenize{#1}\relax
    \expandafter\@firstoftwo
  \else
    \expandafter\@secondoftwo
  \fi
  {\begin{theorem}}{\begin{theorem}[{#1}]}%
}{\end{theorem}}
\newenvironment{corn}[2][]{%
  \renewcommand{\thecorollary}{#2}%
  \if\relax\detokenize{#1}\relax
    \expandafter\@firstoftwo
  \else
    \expandafter\@secondoftwo
  \fi
  {\begin{corollary}}{\begin{corollary}[{#1}]}%
}{\end{corollary}}
\newenvironment{lemn}[2][]{%
  \renewcommand{\thelemma}{#2}%
  \if\relax\detokenize{#1}\relax
    \expandafter\@firstoftwo
  \else
    \expandafter\@secondoftwo
  \fi
  {\begin{lemma}}{\begin{lemma}[{#1}]}%
}{\end{lemma}}
\makeatother
\theoremstyle{plain}
\newtheorem*{uproposition}{Proposition}
\newtheorem*{ulemma}{Lemma}
\newtheorem*{attainmentlemma}{Attainment Lemma}

\DeclareMathOperator*{\argmax}{arg\,max}
\newcommand{\E}{\mathbb{E}}
\newcommand{\Var}{\operatorname{Var}}
\newcommand{\Cov}{\operatorname{Cov}}
\newcommand{\Bias}{\operatorname{Bias}}
\newcommand{\ME}{\mathrm{ME}}
\newcommand{\CV}{\mathrm{CV}}

\begin{document}

\title{Sharp Bounds on the Winner's Bias under Independent and Nonnegatively Correlated Noise}

\author{%
  Ivan Viakhirev\textsuperscript{1}\qquad
  Kirill Borodin\textsuperscript{1,2}\qquad
  Grach Mkrtchian\textsuperscript{1}\\[1.5ex]
  \normalsize\textsuperscript{1}lab260 (\href{https://lab260.ru}{lab260.ru})\qquad
  \textsuperscript{2}BitmanagerAI (\href{https://bitmanager.ai}{bitmanager.ai})\\[1ex]
  \normalsize Correspondence: Kirill Borodin,
  \href{mailto:kborodin.research@gmail.com}{\texttt{kborodin.research@gmail.com}}}

\date{}

\maketitle

\begin{abstract}
Reporting the best of $M$ noisy, unbiased scores overstates the winner's value by up to
$c\sqrt{V}$, where $V$ is the total noise variance and the sharp constant $c$ depends on how the
noise is dependent. It was known only at the two extremes, arbitrary dependence
($\sqrt{(M-1)/M}$) and i.i.d.\ noise. We determine it for two classes in between: for
independent, not identically distributed noise it is $\sqrt{(M-1)/(2M)}$, and for pairwise
nonnegatively correlated noise it is $(M-1)/M$; for uncorrelated noise this answers the maximum
version of an open question about the expected range. Explicit finite constructions attain both.
As a corollary, the bias of the cross-validation estimator is bounded sharply within each class.
All theorems are formally verified in Lean~4. Across 448
language-model and classifier selection pipelines, designs with per-candidate items realize a
median 0.75 of the independent-class ceiling. At a fixed item budget, scoring all candidates on shared items
is markedly less optimistic than giving each candidate its own items.
\end{abstract}

\section{Introduction}\label{sec:intro}

\paragraph{The Problem.}
Selecting the best of $M$ alternatives by noisy scores and reporting the winner's score is a basic
statistical operation. If the selection scores are unbiased with total variance $V$, the reported
maximum is optimistically biased, and the worst case of that bias over noise laws is governed by
\[
c_M \;=\; \sup \frac{\E\bigl[\max_{i \le M} Z_i\bigr]}{\sqrt{\sum_i \Var Z_i}}
\qquad \text{over mean-zero } Z.
\]

\paragraph{What Is Known.}
Under arbitrary dependence the sharp constant is $\sqrt{(M-1)/M}$ \citep{aven1985}, with tightness
in the equal-moments case due to \citet{arnoldgroeneveld1979}. For i.i.d.\ noise the sharp
constant is $(M-1)/\sqrt{M(2M-1)}$ \citep{gumbel1954,hartleydavid1954}. Between these poles,
\citet{bnt2006} compute tight bounds under covariance information as a per-instance semidefinite
program, and \citet{papadatos2014} proves the tight covariance-informed bound at $M = 2$, stating
the general-$M$ closed-form problem as open. The independent, non-identically distributed case
has, to our knowledge, no published sharp constant.

\paragraph{The Gap.}
Neither pole describes how a selection is usually evaluated. Candidates of different accuracy have
different score variances, so their noise is not identically distributed and the i.i.d.\ constant
does not apply. The arbitrary-dependence constant does apply, but it ignores the structure that the
evaluation design fixes. Giving each candidate its own test items makes the noise independent, and
scoring all candidates on the same items typically makes it positively correlated. Both classes sit
between the poles, and for neither is a closed-form sharp constant known. This paper
supplies both.

\paragraph{Contributions.}
\begin{enumerate}
\item \textbf{Independent noise} (main result, Theorem~3). The sharp constant is
  $\sqrt{(M-1)/(2M)}$, attained exactly by one deterministic alternative plus an explicit
  two-point cascade. The proof reduces general laws to two-point laws per arm, splits the problem
  into $M$ convex problems, and closes each with a factored cubic.
\item \textbf{Nonnegative correlation} (Corollary~1b). The sharp constant is $(M-1)/M$, attained
  at finite parameters. For uncorrelated noise this settles the maximum analogue of an open
  question of \citet[Sec.~7]{papadatos2014}. With the two poles, the constants form a graded family,
  strictly ordered for $M \ge 3$ (Table~\ref{tab:graded}).
\item \textbf{Consequences.} The bias of the cross-validation estimator is bounded sharply in each
  class (Corollary~2). With constant one this bound is Conjecture~1 of \citet{vanhasselt2013},
  which already follows from \citet{aven1985}; our constants sharpen it. Against Gaussian noise
  the worst case is close at small $M$ (19\% above for the independent class at $M = 4$) and
  conservative by a factor of order $\sqrt{M/\log M}$ at large $M$ (Table~\ref{tab:gaussian}).
\item \textbf{Selection pipelines} (Sec.~\ref{sec:numerical}). Over 448 configurations of
  language-model and classifier selection, per-candidate designs realize a median 0.75 of the
  independent constant, and the $\Cov \ge 0$ hypothesis fails in one of nine populations, so it
  must be checked on the data. At a fixed item budget, shared items incur a median 4.9 times less
  optimism than per-candidate items, against the factor 2.1 the ceilings predict.
\end{enumerate}

\paragraph{Method Summary.}
Both dependent classes follow from one master inequality: Aven's bound applied to deviations from
the mean, proved by Samuelson's inequality pointwise and then Cauchy--Schwarz
\citep{samuelson1968}. Its covariance term is what separates the classes. The independent case
needs different machinery and occupies most of the appendix.

\paragraph{Formal Verification.}
Every theorem, corollary and lemma in the paper is machine-checked in Lean~4 with Mathlib, in about
7{,}900 lines with no \texttt{sorry} and no axioms beyond Lean's standard three. The statements are
formalized for random variables on an arbitrary probability space, with independence, square
integrability and zero mean as explicit hypotheses, so the formal theorems are those stated here
and not special cases. Attainment and the suprema are formalized too, so sharpness is checked as
well as the bounds. Appendix~\ref{app:lean} maps each result to its Lean name. The formalization
will be released together with the code and the score matrices of the numerical study.

\section{Setup and the Transfer Lemma}\label{sec:setup}

Alternatives (arms) $i = 1, \dots, M$ have true means $\mu_i$, $\mu^* = \max_i \mu_i$, gaps
$g_i = \mu^* - \mu_i \ge 0$. Selection scores are $\hat a_i = \mu_i + Z_i$ with $\E[Z_i] = 0$,
$\Var Z_i = v_i$, $V = \sum_i v_i$. The winner is $I = \argmax_i \hat a_i$ (ties broken
uniformly). Two estimators of $\mu^*$:
\begin{itemize}
\item the \emph{max-estimator} reports $\hat a_I$, with
  $\Bias(\ME) = \E[\hat a_I] - \mu^* \in [0, \E[\max_i Z_i]]$;
\item the \emph{cross-validation} (double) \emph{estimator} evaluates an independent unbiased
  estimate at $I$, with $\Bias(\CV) = -\E[g_I] \in [-\E[\max_i Z_i], 0]$.
\end{itemize}

\begin{lemn}[transfer]{1}
$\E[g_I] \le \E[\max_j Z_j]$.
\end{lemn}

The max-estimator side needs no lemma: $\hat a_I - \mu^* \le Z_I \le \max_j Z_j$ because
$\mu_I \le \mu^*$. So every upper bound on $\E[\max_i Z_i]/\sqrt{V}$ bounds $\Bias(\ME)$ and
$|\Bias(\CV)|$ alike, and
the extremal constructions below approach their constants in the $\E[g_I]$ form as well, in the limit
where each alternative's deficit rises to its largest noise lead over the best alternative
(Appendix~\ref{app:d5}). For $K$-fold cross-validation the bound
applies fold-wise and averages (Appendix~\ref{app:notation}).

\section{Main Results}\label{sec:main}

\begin{thmn}[master bound]{1}\label{thm:master}
For any mean-zero $Z$, with $\bar Z = (1/M)\sum_i Z_i$:
\[
\E\bigl[\max_i Z_i\bigr] \;\le\; \sqrt{(M-1)/M} \cdot \sqrt{V - M\,\E[\bar Z^2]}.
\]
\end{thmn}

This is Aven's (1985) bound applied to the deviations $Z_i - \bar Z$; we record it with a short
self-contained proof (Samuelson's inequality pointwise, then Cauchy--Schwarz;
Appendix~\ref{app:thm1}) because the covariance correction term $M\,\E[\bar Z^2]$ is what
separates the dependence classes below. At $M = 2$ the bound is known to be tight under
covariance information \citep[Cor.~7.1]{papadatos2014}. A quantitative comparison with Aven's
displayed covariance-aware bounds, which run through a fixed reference arm, is given in
Appendix~\ref{app:thm1}: the centered form is tighter by a factor between 1 and $\sqrt{2}$, both
ends attained.

\begin{corn}[arbitrary dependence]{1a}
$\E[\max_i Z_i] \le \sqrt{(M-1)/M} \cdot \sqrt{V}$, sharp \citep{aven1985}; the attaining law is
the uniform-index one-up construction of \citet{arnoldgroeneveld1979}.
\end{corn}

\begin{corn}[pairwise nonnegative correlation; new]{1b}
If $\Cov(Z_i, Z_j) \ge 0$ for all $i \ne j$, then
\[
\E\bigl[\max_i Z_i\bigr] \;\le\; \frac{M-1}{M} \cdot \sqrt{V},
\]
and the constant is sharp already in the uncorrelated subclass $\Cov = 0$, attained there exactly
at finite parameters by an explicit law (uniform-index one-up plus an independent common shock
whose variance cancels the one-up covariances; Appendix~\ref{app:cor1b}). With every covariance
strictly positive the bound is not attained, so the constant is the sharp one for the class and not
for each configuration in it.
\end{corn}

\citet[Sec.~2.4]{bnt2006} compute the tight covariance-informed bound as a per-instance
semidefinite program; Corollary~1b is the closed form of its value on the uncorrelated class, and
answers the uncorrelated case of the question posed by \citet[Sec.~7]{papadatos2014}.

\begin{thmn}[independent noise, deterministic best alternative; new]{2}\label{thm:two}
For independent mean-zero $Z_1, \dots, Z_m$, with $V = \sum_{i=1}^{m} \Var Z_i$ over these $m$
arms:
\[
\E\bigl[\max(0, Z_1, \dots, Z_m)\bigr] \;\le\; \sqrt{m/(2(m+1))} \cdot \sqrt{V},
\]
sharp, attained exactly by the cascade law in which alternative $k$ takes value $k/m$ with
probability $1/(k+1)$ and value $-1/m$ otherwise (Appendix~\ref{app:thm2}).
\end{thmn}

\begin{thmn}[independent noise; main result]{3}\label{thm:three}
For independent mean-zero $Z_1, \dots, Z_M$:
\[
\sup \E\bigl[\max_i Z_i\bigr] / \sqrt{V} \;=\; \sqrt{(M-1)/(2M)},
\]
and the supremum is attained exactly by the $(M-1)$-arm cascade of Theorem~2 together with one
deterministic arm. The optimal role of one alternative is to be a deterministic floor; at $M = 2$
a deterministic arm is necessary for equality (Appendix~\ref{app:d5}).
\end{thmn}

\emph{Proof structure} (Appendix~\ref{app:thm3}): (i) a Dinkelbach-type conjugation shows the
supremum is determined by laws with at most two atoms per arm; the measure-theoretic step is
Corollary~2 of \citet{bayerteichmann2006}, applied to the moment map
$\mu \mapsto (\int d\mu, \int z\, d\mu, \int (g - \lambda z^2)\, d\mu)$, followed by a rank-2
linear program whose optimal vertices have at most two atoms; (ii) in two-point form the
objective is a maximum of $M$ linear functionals of the reweighted atoms $w_i = x_i u_i$, so the
constrained
supremum is a maximum of $M$ explicit values; (iii) each value is bounded by a recursion whose
fixed points are $G_m = m/(2(m+1))$; the one nontrivial branch reduces to the positivity of a
factored cubic on an interval, with equality exactly at the attaining configuration.

\begin{table*}[t]
\centering
\caption{The graded family of sharp constants $\sup \E[\max_i Z_i]/\sqrt{V}$ by dependence
class. ``Exact'' means the supremum is attained at finite parameters.}\label{tab:graded}
\small
\begin{tabular}{llll}
\toprule
dependence class & sharp constant & first proved & attainment \\
\midrule
arbitrary & $\sqrt{(M-1)/M}$ & Aven 1985 / Arnold--Groeneveld 1979 & exact \\
pairwise $\Cov \ge 0$ & $(M-1)/M$ & this paper & exact \\
independent & $\sqrt{(M-1)/(2M)}$ & this paper & exact \\
i.i.d. & $(M-1)/\sqrt{M(2M-1)}$ & Gumbel 1954 / Hartley--David 1954 & exact \\
\bottomrule
\end{tabular}
\end{table*}

\begin{table*}[t]
\centering
\caption{Worst-case constants against the Gaussian value: the last column is
$\E[\max_i Z_i]/\sqrt{V}$ for i.i.d.\ standard Gaussian noise, the same comparator in every row; the
other columns are the sharp constants of Table~\ref{tab:graded}. Entries are rounded to three
decimals, and the percentages quoted in the text are computed from the exact
values.}\label{tab:gaussian}
\begin{tabular}{rccccc}
\toprule
$M$ & arbitrary & $\Cov \ge 0$ & independent & i.i.d.\ worst case & Gaussian i.i.d. \\
\midrule
2 & 0.707 & 0.500 & 0.500 & 0.408 & 0.399 \\
3 & 0.816 & 0.667 & 0.577 & 0.516 & 0.489 \\
4 & 0.866 & 0.750 & 0.612 & 0.567 & 0.515 \\
5 & 0.894 & 0.800 & 0.632 & 0.596 & 0.520 \\
10 & 0.949 & 0.900 & 0.671 & 0.653 & 0.487 \\
20 & 0.975 & 0.950 & 0.689 & 0.680 & 0.418 \\
50 & 0.990 & 0.980 & 0.700 & 0.696 & 0.318 \\
\bottomrule
\end{tabular}
\end{table*}

Table~\ref{tab:graded} collects the graded family. The family is strictly decreasing down the
table for every $M \ge 3$. Two identities: the $\Cov \ge 0$ constant is the square of the
arbitrary one, and the arbitrary-to-independent ratio is $\sqrt{2}$ for every $M$. The
$\Cov \ge 0$ and independent constants coincide only at $M = 2$; the arbitrary constant coincides
with neither.

Table~\ref{tab:gaussian} compares the worst case with the Gaussian value of the same ratio. How far
a constant sits above that value depends on which class it belongs to. The i.i.d.\ constant exceeds
it by 2.3\%, 5.7\%, 10.2\% and 14.6\% at $M = 2, \dots, 5$, so in small comparisons the extremal
i.i.d.\ law is barely worse than a Gaussian one. The independent constant, which is the one
Theorem~\ref{thm:three} supplies and which drops the identical-distribution requirement, sits
further above, and not monotonically: 25.3\%, 18.2\%, 19.0\% and 21.6\% at the same $M$, a minimum
at $M = 3$ and growth after it.

The comparator is i.i.d.\ Gaussian noise in every row; the supremum
over \emph{independent} Gaussian laws of the same total variance is at least that value, so these
percentages bound the independent class's distance from its own Gaussian worst case from above. For large $M$ the Gaussian
ratio
decays like $\sqrt{2 \log M / M}$ while the worst-case constants approach their limits, so both are
conservative by a factor of order $\sqrt{M/\log M}$; conclusions that depend on the difference
between dependence classes at large $M$ should not be drawn from worst-case constants alone.

\begin{figure*}[t]
\centering
\includegraphics[width=\textwidth]{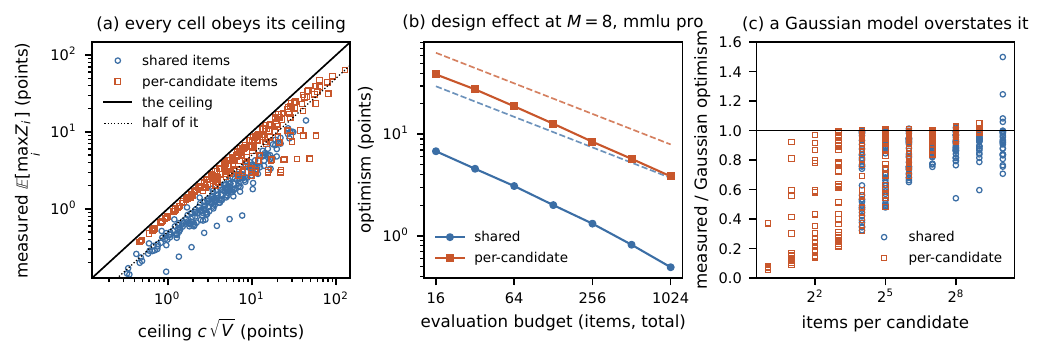}
\caption{Configuration cells of Sec.~\ref{sec:numerical}: one marker per population, pool size
$M$, item budget and design; scores in accuracy points.
\textbf{(a)} Measured $\E[\max_i Z_i]$ against the class ceiling $c\sqrt{V}$, log--log. Solid: the ceiling,
which no marker exceeds; dotted: half of it.
\textbf{(b)} Optimism against total item budget $n$ at $M = 8$ on MMLU-Pro, log--log; colour is the design,
dashed lines are the matching ceilings. Per-candidate items ($n/8$ each) lie above shared items
throughout.
\textbf{(c)} Measured optimism over that of a Gaussian model with matched covariance, against items
per candidate ($n$ shared, $n/M$ per-candidate; log scale); below 1, the Gaussian model overstates. Only cells
resolved to three standard errors are shown.}\label{fig:cells}
\end{figure*}

\section{The Bias of the Cross-Validation Estimator}\label{sec:vh}

\begin{corn}{2}\label{cor:cv}
For the $K$-fold cross-validation estimator of Sec.~\ref{sec:setup},
\[
|\Bias(\CV)| \;\le\; c \cdot \frac{1}{K}\sum_k \sqrt{\textstyle\sum_i \Var \hat a_i^k},
\]
where $c = \sqrt{(M-1)/M}$ in general, $c = (M-1)/M$ under pairwise nonnegative correlation of
the fold's selection scores, and $c = \sqrt{(M-1)/(2M)}$ under independence; each constant is
sharp within its class as the supremum of $\E[\max_i Z_i]/\sqrt{V}$; in the $\Bias(\CV)$ form it is
approached along the limiting configurations of Appendix~\ref{app:d5} rather than attained. The same
three constants bound $\Bias(\ME)$ fold-wise, Lemma~1 covering the cross-validation side and
$\mu_I \le \mu^*$ the other.
\end{corn}

The general case follows from Lemma~1 and Corollary~1a applied fold-wise. Since
$\sqrt{(M-1)/M} < 1$, this settles Conjecture~1 of \citet{vanhasselt2013}, which asserts the
bound with $c = 1$; Corollary~1a is Aven's bound, which appears there as equation~(7). The strict form of
the inequality fails only when every fold variance vanishes, and there both sides are zero.

The $M = 2$ case proved in that paper substitutes $\Var \hat a_1 + \Var \hat a_2$ for
$\Var(\hat a_2 - \hat a_1)$ inside a Cantelli bound. That substitution is valid under
$\Cov \ge 0$, so the constant $1/2$ obtained there is the $\Cov \ge 0$ value at $M = 2$, where
the two classes coincide. They separate for every $M \ge 3$, which is where the graded
constants of Table~\ref{tab:graded} matter.

Which of the three constants applies is decided by the fold, and for standard $K$-fold
cross-validation it is usually not the independent one: within a fold every candidate is scored on
the same held-out block, so the selection scores are dependent and typically positively correlated,
which puts the fold in the $\Cov \ge 0$ class with $c = (M-1)/M$. The independent constant governs
the case where each candidate gets its own evaluation data, which is the per-candidate design of
Sec.~\ref{sec:numerical}.

A remark in the same paper suggests that i.i.d.\ noise is the worst case within the independent
class. Table~\ref{tab:graded} shows otherwise: the i.i.d.\ constant lies strictly below the
independent one for every $M \ge 2$, so the i.i.d.\ configuration does not govern the
independent class.

\section{Numerical Study: Selection Pipelines}\label{sec:numerical}

\begin{table*}[t]
\centering
\caption{Optimism and regret of best-of-$M$ selection at $M = 8$ with a fixed budget of 64 items,
under both evaluation designs: shared items ($\Cov \ge 0$ class, 64 items per candidate) and
per-candidate items (independent class, 8 items each; ``per-cand.'' in the table). Scores are percentage points of accuracy;
20{,}000 replications per cell. Optimism is $\E[\hat a_I] - \mu^*$, regret is $\mu^* - \E[\mu_I]$,
the ceiling is $c\sqrt{V}$ for the design's class, $\rho$ is the fraction of that ceiling realized
by the measured $\E[\max_i Z_i]$, and the last column is the optimism a Gaussian model with the
matched covariance predicts.}\label{tab:cells}
\small
\setlength{\tabcolsep}{4pt}
\begin{tabular}{lrlrr@{\,$\pm$\,}lrrrrr}
\toprule
population & $N$ & design & items & \multicolumn{2}{c}{optimism} & regret & $\E[\max_i Z_i]$ &
ceiling & $\rho$ & Gaussian \\
\midrule
MMLU-Pro          & 4000 & shared   &  64 &  3.09 & 0.04 & 0.44 &  5.15 & 14.86 & 0.346 &  3.05 \\
MMLU-Pro          & 4000 & per-cand. &   8 & 18.87 & 0.07 & 2.54 & 24.07 & 31.78 & 0.757 & 19.02 \\
GSM8K             & 1319 & shared   &  64 &  1.09 & 0.02 & 0.48 &  4.27 & 10.92 & 0.391 &  1.17 \\
GSM8K             & 1319 & per-cand. &   8 &  6.87 & 0.02 & 0.45 & 17.59 & 23.34 & 0.754 &  8.90 \\
ARC-Challenge     & 1172 & shared   &  64 &  1.75 & 0.02 & 0.42 &  2.54 &  8.81 & 0.288 &  1.87 \\
ARC-Challenge     & 1172 & per-cand. &   8 &  8.18 & 0.01 & 0.30 & 10.05 & 18.84 & 0.534 & 13.13 \\
MATH-500          &  500 & shared   &  64 &  1.19 & 0.04 & 0.92 &  4.80 & 13.38 & 0.359 &  1.19 \\
MATH-500          &  500 & per-cand. &   8 & 13.71 & 0.07 & 1.39 & 21.73 & 28.61 & 0.760 & 14.27 \\
digits (hard)     & 1757 & shared   &  64 &  1.91 & 0.04 & 0.54 &  4.23 & 13.96 & 0.303 &  2.02 \\
digits (hard)     & 1757 & per-cand. &   8 & 17.31 & 0.05 & 1.37 & 21.17 & 29.84 & 0.709 & 19.31 \\
digits (sweep)    &  997 & shared   &  64 &  0.78 & 0.01 & 0.34 &  2.12 &  5.56 & 0.381 &  1.14 \\
digits (sweep)    &  997 & per-cand. &   8 &  2.11 & 0.00 & 0.06 &  4.25 & 11.89 & 0.358 &  7.71 \\
digits (16 seeds) &  997 & shared   &  64 &  0.82 & 0.01 & 0.31 &  1.62 &  5.17 & 0.314 &  1.16 \\
digits (16 seeds) &  997 & per-cand. &   8 &  2.11 & 0.00 & 0.10 &  3.16 & 11.04 & 0.286 &  7.58 \\
breast cancer     &  369 & shared   &  64 &  1.02 & 0.01 & 0.16 &  1.47 &  5.11 & 0.288 &  1.35 \\
breast cancer     &  369 & per-cand. &   8 &  2.44 & 0.00 & 0.05 &  2.98 & 10.93 & 0.273 &  7.90 \\
wine              &  118 & shared   &  64 &  1.02 & 0.01 & 0.60 &  2.70 &  6.98 & 0.387 &  1.22 \\
wine              &  118 & per-cand. &   8 &  3.39 & 0.00 & 0.23 &  7.02 & 14.93 & 0.470 &  9.25 \\
\bottomrule
\end{tabular}
\end{table*}

\paragraph{Data and Pipelines.}
Two families of selection problems. The first is language-model configuration selection: one
instruction-tuned 7B model (Qwen2.5-7B-Instruct; \citealp{qwen2024qwen25}) crossed with decoding
temperature $0$ and $0.7$, zero-shot and five-shot prompting, and chain-of-thought on and off,
giving $M = 8$ candidates evaluated item by item on four benchmarks --- MMLU-Pro
\citep[a 4000-item uniform subsample of 12032;][]{wang2024mmlupro}, GSM8K
\citep[1319 items;][]{cobbe2021gsm8k}, ARC-Challenge \citep[1172 items;][]{clark2018arc} and
MATH-500 \citep[500 items;][]{hendrycks2021math,lightman2024verify}. An answer-extraction grader
marks each generation right or wrong, so an item score is in $\{0,1\}$ for the greedy candidates and in
$\{0, \tfrac12, 1\}$ for the temperature-0.7 candidates, which are drawn twice per item and
averaged. The averaging halves those candidates' per-item variance and is part of the variance
heterogeneity measured below.

The second is classifier selection: four model families (regularized logistic regression,
$k$-nearest neighbours, decision trees, small multilayer perceptrons) at four capacities each,
giving $M = 16$ candidates on handwritten digits \citep{ucidigits} at two difficulty levels,
breast cancer \citep{ucibreast} and wine \citep{uciwine}, plus a pool of 16 seeds of one fixed
configuration. All three datasets are the copies shipped with scikit-learn
\citep{pedregosa2011sklearn}.

\paragraph{Designs and Definitions.}
Each candidate is scored once on every item of its population, so the item mean of row $i$ is that
candidate's population score $\mu_i$ and no candidate is ever retrained. Evaluation designs are then
realized by resampling items: the \emph{shared} design scores every candidate on the same $n$ items,
the \emph{per-candidate} design draws an independent set of $n/M$ items for each candidate. Both
draw i.i.d.\ with replacement from the population, so the scoring noise has exactly zero mean; the
per-candidate sets are drawn independently of one another rather than partitioned, so they may
overlap in content, and it is the independence of the draws, not disjointness, that makes the
per-candidate noise exactly independent across candidates.

We report the \emph{optimism} of the
reported best, $\E[\hat a_I] - \mu^*$, which is $\Bias(\ME)$ of Sec.~\ref{sec:setup} on the accuracy
scale, and its \emph{regret} $\mu^* - \E[\mu_I]$. The \emph{ceiling} of a configuration is
$c\sqrt{V}$ for the constant $c$ of its class, and which class that is follows from the design: the
per-candidate design is in the independent class by construction, whatever the population's
covariances, so the $\Cov \ge 0$ test below is applied to the shared design alone.

\paragraph{Grid.}
One \emph{cell} is one population, pool size, budget and design. A pool of size $M$ is the first $M$
candidates of its population, so the pools are nested in $M$. Candidates are ordered by the sweep
that generates them, so the small pools vary only the fastest axis and are correspondingly
homogeneous: at $M = 2$ a language-model pool is one prompting scheme at two temperatures and a
classifier pool is two capacities of one model family. Results at $M = 2$ and $4$ should be read
with that in mind.

The language-model populations have eight
candidates and the classifier populations sixteen, so pool sizes run over $M \in \{2, 4, 8\}$ and
$M \in \{2, 4, 8, 16\}$ respectively. Budgets double, $n \in \{16, 32, \dots, 1024\}$, and each cell
is replicated 20{,}000 times, giving $(4 \cdot 3 + 5 \cdot 4) \cdot 7 \cdot 2 = 448$ cells. Scores
below are percentage points.

\paragraph{Validity of the Ceiling.}
In every cell the measured $\E[\max_i Z_i]$ lies on or below the ceiling of its class
(Figure~\ref{fig:cells}(a)). Since each design's class is fixed by construction, a violation would
mean a fault in the measurement rather than a counterexample, and we report the check in that
spirit. It does exercise the extremes: cells in which the per-candidate design leaves one item per
candidate, and the per-candidate cells of the pool that fails the covariance test, which are held to
the independent constant rather than to the looser one its population would suggest.

\paragraph{Realized Fraction.}
The realized fraction $\rho = \E[\max_i Z_i]/(c\sqrt{V})$ separates the two classes: with
per-candidate items (independent class, Theorem~3) $\rho$ has median 0.75 and reaches 0.84, and with
shared items it has median 0.37 and reaches 0.62, each cell divided by the constant of its own
class. These medians run over all pool sizes and budgets; Table~\ref{tab:cells} is one budget, and
its nine per-candidate rows have a median of 0.53, because $\rho$ rises with the budget.

The
ceiling of Theorem~3 is therefore about a third above the $\E[\max_i Z_i]$ an
independent-evaluation pipeline actually incurs. The optimism of the reported best is smaller
again, so the ceiling exceeds it by more; the reporting recipe below quotes both distances.

The $\Cov \ge 0$ ceiling is the looser of the two because positive correlation across candidates
cancels in the score differences that decide the selection without cancelling in $\sqrt{V}$, so
the denominator grows while the numerator does not.

Where a pool falls inside its class depends on how close its candidates are to the top of the
accuracy scale. At $M = 8$ with eight items per candidate (Table~\ref{tab:cells}) the
independent-class fraction is 0.71 to 0.76 on the four pools that still contain a candidate below
0.75 --- MMLU-Pro, GSM8K, MATH-500 and the hardest classifier pool --- and falls as the weakest
candidate rises: 0.53 on ARC-Challenge (0.895 to 0.917), 0.47 on wine (0.924 to 0.966), and 0.27 to
0.36 on the three pools whose candidates all exceed 0.95.

Near the top of the scale the per-item
variance, at most $p(1-p)$, collapses, and a mean of eight bounded items is far from the two-point
law that attains the constant. This ordering is a regularity of nine pools at one budget, not a law; which
feature of a pool predicts $\rho$ is open (Sec.~\ref{sec:limitations}).

\paragraph{Verifying the Class Hypothesis.}
All four language-model populations and four of the five classifier pools have every pairwise
covariance of per-item scores positive, so Corollary~1b applies and the constant is $(M-1)/M$. The
16-candidate wine pool contains a negatively correlated pair (minimum covariance $-0.031$), which
places it in the arbitrary-dependence class and raises its ceiling from $(M-1)/M = 0.938$ to
$\sqrt{(M-1)/M} = 0.968$; its eight-candidate shortlist has all covariances positive. The graded
family is applied by checking the covariances and reading off the constant.

\paragraph{Design Effect at a Fixed Budget.}
Splitting a fixed budget into $M$ blocks multiplies each candidate's variance by $M$, so the
$\sqrt{2}$ that independence buys in the constant is a statement per unit of variance rather than
design advice. At equal budget the ceilings predict a ratio $M/\sqrt{2(M-1)}$ of per-candidate to
shared optimism, between 1.41 and 2.92 over our pool sizes; the measured ratio has median 4.85 with
interquartile range 3.14 to 8.63, and the shared design incurs less optimism in every one of the 220
pairs in which both quantities are resolved to three standard errors (Figure~\ref{fig:cells}(b)).
The measured effect exceeds the predicted one because the shared design also realizes a smaller
fraction of its own ceiling.

In the MMLU-Pro cell of Table~\ref{tab:cells} the same 64 items cost
3.09 points of optimism when shared and 18.87 points when split into eight blocks of eight, a factor
6.1. Shared evaluation dominates at a fixed budget, and by more than the constants alone suggest.

\paragraph{Comparison with a Gaussian Model.}
A Gaussian model with the matched covariance is not a substitute for the ceiling. Replacing the
resampled scores by Gaussian ones changes the optimism by a factor whose median is 0.89
(interquartile range 0.74 to 0.97 over 444 resolved cells), and the error is systematic in the
budget, the median rising monotonically from 0.60 at $n = 16$ to 0.96 at $n = 1024$.

At small
budgets the Gaussian model \emph{overstates} the optimism, by up to sixteenfold, because an average
of a few bounded discrete item scores has much lighter tails than its Gaussian counterpart: on the
three pools whose eight candidates all exceed 0.95 --- the two digit sweeps and breast cancer --- it
predicts 7.6 to 7.9 points where 2.1 to 2.4 are incurred (Table~\ref{tab:cells}).

The direction
reverses in 24 of 444 cells, by up to 1.5-fold at the largest budgets and extreme pass rates, where
skewness dominates (Figure~\ref{fig:cells}(c)); a handful of further cells cross the line by a
margin of no consequence once the ratio is already near one, as the shared MMLU-Pro row does at a
budget of 64 with 3.09 against 3.05.

\paragraph{Variance Heterogeneity.}
The extremal laws that attain Theorem~3 carry variances spanning a factor of order $M$ across arms.
Real pools do not: the ratio of largest to smallest per-candidate item variance stays between 1.15
and 1.81 across our nine populations, and within that band $\rho$ does not track it, the Spearman
rank correlation being 0.25. The accuracy level does track it, at $-0.95$ over the same nine
populations. Nine points are few either way, and the contrast is what we report rather than a
predictor: one ordering is monotone up to an adjacent swap and spans a threefold range in $\rho$,
the other moves without order inside a narrow band.

Attainment needs a variance profile no selection
pool we measured comes close to, which is why the constants are best read as ceilings computable
from data, not as predictions of the optimism.

\paragraph{A Worked Cell.}
Selecting among the eight configurations by MMLU-Pro accuracy measured on 64 shared items, the
accuracy of the reported winner overstates its population accuracy by $3.09 \pm 0.04$ points against
a ceiling of $14.86$ points, while the winner's population accuracy is $0.44$ points below the best
available. The first number is what a leaderboard reports; the second is what the selection
costs.

The seed pool of Table~\ref{tab:cells} is the closest thing here to a null case: sixteen
seeds of one configuration, whose population accuracies still differ, by 2.4 points over the sixteen
and 1.1 over the eight in the table, so most but not all of the reported gain of the best seed is
winner's bias. Its measured quantities behave as in the other pools, as the theory implies, since
the constants do not depend on the means.

\paragraph{Reporting Recipe.}
The optimism of a leaderboard-style best-of-$M$ number admits a ceiling computable from quantities
already in hand: the per-candidate score variances give $\sqrt{V}$, and the class constant follows
from the evaluation design, after checking the pairwise covariances if the design is shared. No
distributional assumption and no additional model runs are required.

Against $\E[\max_i Z_i]$, the
quantity the theorem bounds, that ceiling is too large by a factor between 1.3 and 1.7 over the
interquartile range of cells with per-candidate items, and between 2.4 and 3.2 with a shared
evaluation set; against the optimism of the reported best, which is smaller again,
the same ranges are 1.5 to 3.3 and 4.5 to 11.7. A recipe must say which of the two it bounds, since the optimism of the reported best is
smaller than $\E[\max_i Z_i]$ whenever the candidates are not tied.

\paragraph{Exclusions.}
A 60-item competition benchmark (AIME 2024 and 2025) was evaluated and excluded: at this model scale
four of the eight candidates score exactly zero, so the pool has four identical rows with no
variance and selection has nothing to select between. On MATH-500 and MMLU-Pro between 5\% and 20\%
of chain-of-thought generations reach the token cap, which lowers the population accuracies of the
affected candidates; since the population is defined by the measured scores and the optimism is
measured against that same population, this does not affect any statement above, but it does mean
the absolute accuracies are not comparable with published leaderboard numbers.

\section{Related Work}\label{sec:related}

\paragraph{Classical Moment Bounds.}
\citet{aven1985} proves
\[
\E[\max_i X_i] \le \max_i \E X_i + \sqrt{(n-1)/n \cdot \textstyle\sum_i \Var X_i},
\]
in which $n$ is the
number of variables, $M$ in our notation, and gives covariance-aware variants through a fixed
reference arm. \citet{arnoldgroeneveld1979} prove tightness in the equal-moments case with an
explicit dependent construction. These are the arbitrary-dependence pole of Table~\ref{tab:graded};
the i.i.d.\ pole is \citet{gumbel1954} and \citet{hartleydavid1954}.

\paragraph{Convex-Optimization Bounds.}
\citet{bnt2006} treat the moment problem for order statistics under arbitrary dependence (``no
other assumptions on independence or the type of distribution are made'') and compute the
covariance-informed tight bound as a semidefinite program; the only independence-adjacent result
they cite is the i.i.d.\ bound above.

In the marginal-moment-problem line that follows them,
expectations of maxima and ranges are bounded by convex optimization under marginal or partial
dependence information, and \citet{natarajanteo2017} reduce the semidefinite programs involved and
apply them to the expected range. Those bounds are optimal values computed per instance, and they
require the covariance information that the independent class replaces by a structural assumption.

\paragraph{Bounds for Samples from a Common Distribution.}
The projection method of \citet{gajekrychlik1996,gajekrychlik1998} and the
monograph of \citet{rychlik2001} derive sharp expectation bounds for order statistics of samples
from a common distribution, possibly restricted in shape (decreasing density, decreasing failure
rate, symmetric unimodal). The independent case treated there \citep{gajekrychlik1998} concerns
independent samples from one such distribution, so the variables are identically distributed and
the bounds are in units of the common mean and standard deviation, which is a different problem
from ours.

\citet{rychlik1995} bounds order statistics of dependent variables with given
non-identical marginals, and \citet{kaluszka2005} bound L-statistics from dependent samples of
random length. None of this work treats independent, non-identically distributed variables under
per-variable variance information, which is the setting of Theorems~2 and~3.

\paragraph{The Open Covariance-Informed Problem.}
\citet{papadatos2014} derives the tight mean-variance bound for the expected range under arbitrary
dependence, proves the tight covariance-informed bound at $M = 2$ (Corollary~7.1, crediting
\citealp{papadatos2001}), and poses tight covariance-informed bounds at general $M$ as open,
listing the bounds of \citet{aven1985} and \citet{papadatos2001} as non-tight. Corollary~1b answers
the uncorrelated case of that question in closed form.

\paragraph{Estimating the Maximum of Means.}
\citet{vanhasselt2010,vanhasselt2013} introduced the double estimator and the conjecture resolved
in Sec.~\ref{sec:vh}. \citet{deramo2016} propose a Gaussian approximation estimator for the same
quantity. This line studies estimators; our results bound the worst case of any selection-based
report.

\paragraph{Selection Bias in Model Evaluation.}
The minimum cross-validated error of a tuned model is optimistic: \citet{varmasimon2006} show the
bias on simulated data and remove it by nested cross-validation, \citet{cawleytalbot2010} find its
effect comparable to the differences between learning algorithms, and \citet{tibshirani2009bias}
estimate it from the cross-validation run itself. These works measure or estimate the bias in a
given problem; the constants here bound it in advance. Reusing one holdout set or leaderboard for
repeated selection raises the same issue, which \citet{dwork2015reusable} and
\citet{blumhardt2015ladder} control algorithmically; \citet{recht2019imagenet} find that the
accuracy drops on new CIFAR-10 and ImageNet test sets are not explained by such adaptive
overfitting.

\paragraph{Post-Selection Inference.}
Post-selection inference corrects selection effects either simultaneously over all possible
selections \citep{berk2013valid} or by conditioning on the selection event
\citep{lee2016exact,fithian2014optimal}, and \citet{andrews2024winners} build estimators and
confidence intervals for the winner among several estimates. This work is complementary: it
requires distributional assumptions and yields valid inference, whereas the constants here are
distribution-free bounds on unconditional bias.

\section{Limitations}\label{sec:limitations}

The constants are worst-case over noise laws within each dependence class;
Table~\ref{tab:gaussian} quantifies the gap to Gaussian noise, which grows with $M$. The
attainment analysis leaves one open question: whether every attaining configuration in Theorem~3
must contain a deterministic arm is proved at $M = 2$ and conjectured for general $M$ (the
conjecture is used nowhere).

The numerical study of Sec.~\ref{sec:numerical} has a narrow scope. The language-model pools vary
the decoding and prompting configuration of a single 7B model, so pools that mix model scales --- the
case whose variance profile could come closest to the extremal one --- are untested, and the
classifier pools are small models on datasets that ship with a standard library.

All scores are
single-item correctness, and both designs are idealizations: the resampling of
Sec.~\ref{sec:numerical} is what makes the scoring noise exactly mean-zero and the per-candidate
design exactly independent. It is also the only source of randomness, since each candidate is scored
once and its population score is treated as fixed, so the measurements say nothing about noise from
retraining, from data splits, or from distribution shift between the evaluation pool and
deployment.

The realized fractions of
Sec.~\ref{sec:numerical} describe these pools at these budgets and are not typical-case guarantees;
outside them we make no claim about magnitudes beyond Table~\ref{tab:gaussian}. Finally, the gap
between the ceilings and the measurements is explained only qualitatively: the variance
heterogeneity the extremal laws require is absent from these pools, and $\rho$ tracks the accuracy
level instead, but nine populations do not yield a quantitative predictor of $\rho$.

\section*{AI Use Statement}

In this work, we used generative AI tools to assist in writing the Lean~4 formalization of our
proofs (Appendix~\ref{app:lean}). We have not
used generative AI tools to develop the theory, formulate the mathematical claims, provide
ingredients for or write the proofs of Appendices~\ref{app:notation} to~\ref{app:thm3}, propose
hypotheses, design the experiments, implement the methods, generate or clean data, or interpret
the results; translation and qualitative or thematic data analysis are not applicable to this
work. Additionally, we used generative AI tools to edit the text for readability, to create or
modify figures, and to brainstorm the title. We have reviewed all AI-assisted work: the authors
checked every statement of the Lean formalization manually against the paper after it was
implemented, the formal proofs are checked by Lean's kernel, and the authors checked all proofs
in the paper line by line. We take responsibility for the final content of this work, including
text, claims, or artifacts produced with the aid of generative AI.

\clearpage
\appendix
\begin{center}
{\LARGE\bfseries Appendix}
\end{center}
\vspace{1ex}

Numbering matches the main text: Lemma~1 (transfer), Theorem~1 with Corollaries~1a/1b,
Theorem~2, Theorem~3, and the Attainment Lemma. Every identity and attainment claim below was
additionally verified in exact rational and symbolic arithmetic; the verification script will be
released with the code. All theorems and lemmas below are also formally verified in Lean~4 with
Mathlib; the formalization is described in Appendix~\ref{app:lean}.

\section{Notation and the Transfer Lemma}\label{app:notation}

\subsection{Setup and Notation}\label{app:setup}

Alternatives (arms) $i = 1, \dots, M$, true means $\mu_i$, $\mu^* = \max_i \mu_i$, gaps
$g_i = \mu^* - \mu_i \ge 0$. Selection estimates $\hat a_i = \mu_i + Z_i$ with $\E[Z_i] = 0$,
$\Var Z_i = v_i$, $V = \sum_i v_i$, $\bar Z = (1/M) \sum_i Z_i$. The winner is
$I = \argmax_i \hat a_i$. Two estimators of $\mu^*$:
\begin{itemize}
\item \textbf{Max-estimator (ME):} report $\hat a_I = \max_i \hat a_i$;
  $\Bias(\ME) = \E[\max_i \hat a_i] - \mu^* \in [0, \E[\max_i Z_i]]$.
\item \textbf{CV / double estimator:} evaluate an independent unbiased estimate at $I$;
  $\Bias(\CV) = -\E[g_I] \in [-\E[\max_i Z_i], 0]$.
\end{itemize}

\paragraph{Tie Convention.}
Ties in $\argmax$ are broken by uniform choice; none of the extremal laws in this appendix
produce ties. The transfer below also holds under the tie convention of \citet{vanhasselt2013},
which averages the gap over the tied set; no ties occur in the constructions.

\paragraph{Two-Point Notation}(used throughout Appendices~\ref{app:thm2} and~\ref{app:thm3}).
A two-point mean-zero arm $i$ takes the up value $u_i > 0$ with probability $x_i$ and the down
value $d_i = -x_i u_i / q_i$ with probability $q_i = 1 - x_i$; its variance is
$v_i = x_i u_i^2 / q_i$. With arms sorted so that $u_1 \ge \dots \ge u_m$:
\begin{align*}
s_i &= \prod_{j<i} q_j \quad \text{(head survival; $s_1 = 1$),}\\
Q &= \prod_j q_j \quad \text{(all-down probability),}\\
P_k &= \prod_{j>k} q_j \quad \text{(tail survival; $P_M = 1$),}\\
w_i &= x_i u_i,\\
f(x) &= f_m(x) = \sum_i x_i q_i s_i^2,\\
G_m &= \frac{m}{2(m+1)}, \quad G_0 = 0,\\
R(c) &= \max_{x \in [0,1]} \bigl[\, x(1-x) + (1-x)^2 c \,\bigr].
\end{align*}

\subsection{Lemma 1 (Transfer)}\label{app:lemma1}

\begin{lemn}{1}
$\E[g_I] \le \E[\max_j Z_j]$.
\end{lemn}

\begin{proof}
Since $I = \argmax_i \hat a_i$, for a fixed optimal arm $i^\circ$ (one with
$\mu_{i^\circ} = \mu^*$) we have $\mu_I + Z_I \ge \mu^* + Z_{i^\circ}$, i.e.\
$g_I \le Z_I - Z_{i^\circ} \le \max_j Z_j - Z_{i^\circ}$. Take expectations and use
$\E[Z_{i^\circ}] = 0$.
\end{proof}

Consequently every upper bound on $\E[\max_i Z_i]/\sqrt{V}$ in this appendix bounds both
$|\Bias(\CV)|$ and $\Bias(\ME)$, and all three extremal constructions approach their constants in
the $\E[g_I]$ form as well, along the limiting configurations of Appendix~\ref{app:d5}: each gap
rises to $(1-\varepsilon)$ times the largest lead its arm's noise can take over the optimal arm,
$\varepsilon \to 0^+$. That lead is the up-value of the arm for the cascade, whose best arm is
deterministic, but $Md$, not the up-value $(M-1)d$, for the one-up laws of Corollaries~1a
and~1b.

\begin{remark}[$K$-fold decomposition]
For $K$-fold CV the per-fold decomposition is licensed by van Hasselt's (2013) observation that
averaging over folds does not change the bias (``The bias of the CV estimators is unaffected by
the fact that $\hat\mu_{\CV}^*$ averages over $K$ estimators''). Per fold $k$, the tie-averaged
gap is bounded by $\E[\max_j Z_j^k] \le \sqrt{(M-1)/M} \cdot \sqrt{\sum_j \Var Z_j^k}$
(Corollary~1a, which requires no dependence assumption at all), and $\sqrt{(M-1)/M} < 1$; the
bound applies fold-wise and averages. Edge case: the conjectured \emph{strict} inequality
$|\Bias| < \sqrt{\sum \Var}$ fails exactly when all fold variances vanish (both sides are then
$0$); this is noted in the main text.
\end{remark}

\section{Theorem 1 and Corollaries 1a/1b}\label{app:thm1}

\subsection{Theorem 1 (Master Bound)}\label{app:thm1statement}

\begin{thmn}{1}
For any mean-zero $Z$ (no independence assumed),
\[
\E\bigl[\max_i Z_i\bigr] \;\le\; \sqrt{(M-1)/M} \cdot \sqrt{V - M\,\E[\bar Z^2]}.
\]
\end{thmn}

\begin{proof}
Samuelson's inequality \citep{samuelson1968}, applied pointwise: for any reals
$a_1, \dots, a_M$ with mean $\bar a$,
\[
\max_i (a_i - \bar a) \;\le\; \sqrt{\frac{M-1}{M} \cdot \sum_j (a_j - \bar a)^2}.
\]
Apply at $a = Z$, so that $\max_i Z_i = \max_i (Z_i - \bar Z) + \bar Z$. Take expectations, use
$\E[\bar Z] = 0$, and apply Cauchy--Schwarz to $\E\sqrt{\sum_j (Z_j - \bar Z)^2}$:
\begin{align*}
\E\bigl[\max_i Z_i\bigr] &= \E\bigl[\max_i (Z_i - \bar Z)\bigr] + \E[\bar Z]\\
&\le \sqrt{(M-1)/M} \cdot \E\sqrt{\textstyle\sum_j (Z_j - \bar Z)^2}\\
&\le \sqrt{(M-1)/M} \cdot \sqrt{\E \textstyle\sum_j (Z_j - \bar Z)^2},
\end{align*}
and the variance identity
\[
\E \sum_j (Z_j - \bar Z)^2 = V - M\,\E[\bar Z^2]
\]
completes the proof.
\end{proof}

\paragraph{Difference Form.}
Via the identity
\[
\sum_{i<j} \E(Z_i - Z_j)^2 = M \cdot (V - M\,\E[\bar Z^2]),
\]
Theorem~1 is equivalent to a bound in the variances of pairwise differences.

\paragraph{Provenance.}
Theorem~1 is Aven's (1985) $u_{20}$ bound applied to the deviations $Y_i = Z_i - \bar Z$:
$\E \max Z = \E \max Y + \E \bar Z = \E \max Y$, $\E[Y_i] = 0$, and
$\sum_i \Var Y_i = \E \sum_i (Z_i - \bar Z)^2 = V - M\,\E[\bar Z^2]$, so $u_{20}(Y)$ is exactly
the master bound. We therefore claim no novelty for Theorem~1 itself; the Samuelson proof above
is kept as self-contained exposition. At $M = 2$ the bound is published with tightness:
\citet[Cor.~7.1]{papadatos2014}, crediting \citet{papadatos2001}.

\begin{ulemma}[gap to Aven's fixed-arm family]
Aven's covariance-aware bounds are (centered) $u_{2j} = \sqrt{(M-1)/M} \cdot \sqrt{S_j}$ with
$S_j = \sum_i \E(Z_i - Z_j)^2$, for each fixed reference arm $j$. With
$D = V - M\,\E[\bar Z^2]$:
\begin{enumerate}
\item $D \le S_j$ for every $j$, with exact slack $S_j - D = M\,\E[(\bar Z - Z_j)^2]$;
\item $\sum_{j=1}^M S_j = 2M \cdot D$, hence $\min_j S_j \le 2D$;
\item so $\min_{1 \le j \le M} u_{2j} / \mathrm{master} =
  \sqrt{\min_{1 \le j \le M} S_j / D} \in [1, \sqrt{2}]$ (assuming $D > 0$); over this
  reference-arm subfamily the ratio equals $1$ iff some $Z_j = \bar Z$ a.s.\ (by the slack in 1;
  a deterministic arm is neither necessary nor sufficient), and $\sqrt{2}$ is attained, for any
  non-degenerate exchangeable law ($\E(Z_i - Z_j)^2 \equiv e > 0$, e.g.\ i.i.d.), where every
  $S_j = (M-1)e = 2D$. Aven's $j = 0$ member $u_{20} = \sqrt{(M-1)/M} \cdot \sqrt{V}$ is excluded
  from this min; including it keeps the ratio in $[1, \sqrt{2}]$ (since $D \le V$) but changes
  the equality case to ``$\bar Z \equiv 0$ a.s.\ or some $Z_j = \bar Z$ a.s.'': e.g.\ the $M = 2$
  anti-correlated pair of Corollary~1a has $u_{20} = \mathrm{master}$ with no $Z_j = \bar Z$.
\end{enumerate}
\end{ulemma}

\begin{proof}
(1) Fix $j$, set $a_i = Z_i - Z_j$; then $\bar a = \bar Z - Z_j$, $Z_i - \bar Z = a_i - \bar a$,
and pointwise $\sum_i (a_i - \bar a)^2 = \sum_i a_i^2 - M \bar a^2$; take expectations.
(2) Expand $\sum_{i,j} (z_i - z_j)^2 = 2M \sum_i z_i^2 - 2 (\sum_i z_i)^2$ and take expectations.
(3) Immediate from (1)--(2).
\end{proof}

\begin{remark}
Under full covariance information the tight bound is the semidefinite program of
\citet[Sec.~2.4]{bnt2006}, which dominates the master bound; the master bound's virtue is closed
form. \citet[Sec.~7]{papadatos2014} poses tight covariance-informed bounds at general $M$ as
open, listing the bounds of \citet{aven1985} and \citet{papadatos2001} as non-tight;
Corollary~1b below answers the uncorrelated case of that question sharply.
\end{remark}

\subsection{Corollary 1a (Arbitrary Dependence)}\label{app:cor1a}

\begin{corn}{1a}
$\E[\bar Z^2] \ge 0$ in Theorem~1 gives
\[
\E\bigl[\max_i Z_i\bigr] \;\le\; \sqrt{(M-1)/M} \cdot \sqrt{V}.
\]
This is Aven's $u_{20}$ bound (displayed as bound (3) in Bertsimas--Natarajan--Teo's
restatement), cited by van Hasselt as his equation (7). It is sharp.
\end{corn}

\paragraph{Equality Characterization.}
Equality in the chain of the proof of Theorem~1 requires all three of:
\begin{enumerate}
\item $\bar Z \equiv 0$ a.s.\ (to drop the $M\,\E[\bar Z^2]$ term),
\item the Samuelson equality pattern pointwise (one value up, the remaining $M-1$ values equal),
  and
\item $\sum_j (Z_j - \bar Z)^2$ a.s.\ constant (the Jensen/Cauchy--Schwarz condition).
\end{enumerate}

\paragraph{Attainment.}
At $M = 2$: the anti-correlated pair $Z_1 = -Z_2 = \pm s/2$ (each sign with probability $1/2$)
has $\bar Z \equiv 0$ and satisfies the Samuelson equality pattern, attaining $\sqrt{1/2}$. The
anti-correlated pair attains the constant at $M = 2$ only; at general $M$ the attaining law is
the \textbf{uniform-index one-up law} (Arnold--Groeneveld's dependent construction; tightness in
the equal-moments case is theirs, as restated by Bertsimas--Natarajan--Teo): let $J$ be uniform
on $\{1, \dots, M\}$ and
\[
\begin{gathered}
Y_i = (M-1) \cdot d \ \text{ if } i = J,\\
Y_i = -d \ \text{ otherwise} \qquad (d > 0).
\end{gathered}
\]
Then $\E[Y_i] = 0$, $\bar Z \equiv 0$ pointwise, each realization has one value up and the rest
equal, $\Var Y_i = (M-1)d^2$, $V = M(M-1)d^2$, and $\max_i Y_i = (M-1)d$ a.s., so
\[
\E\bigl[\max_i Y_i\bigr] / \sqrt{V} = \frac{(M-1)d}{\sqrt{M(M-1)}\,d} = \sqrt{(M-1)/M},
\]
exactly, at finite parameters.

\subsection{Corollary 1b (Pairwise Nonnegative Correlation; Sharp)}\label{app:cor1b}

\begin{corn}{1b}
If $\Cov(Z_i, Z_j) \ge 0$ for all $i \ne j$, then
\[
\E\bigl[\max_i Z_i\bigr] \;\le\; \frac{M-1}{M} \cdot \sqrt{V},
\]
and this is sharp already in the uncorrelated subclass ($\Cov = 0$), attained exactly at finite
parameters.
\end{corn}

\begin{proof}[Proof of the bound]
$\E[\bar Z^2] = (1/M^2) \cdot \bigl( V + \sum_{i \ne j} \Cov(Z_i, Z_j) \bigr)$. Under pairwise
$\Cov \ge 0$ this gives $\E[\bar Z^2] \ge V/M^2$ (with equality iff all pairwise covariances
vanish). Then Theorem~1 gives
\[
\begin{aligned}
\E\bigl[\max_i Z_i\bigr] &\le \sqrt{\tfrac{M-1}{M}} \sqrt{V - \tfrac{V}{M}}\\
&= \sqrt{\tfrac{M-1}{M}} \sqrt{V \cdot \tfrac{M-1}{M}} = \tfrac{M-1}{M} \sqrt{V}. \qedhere
\end{aligned}
\]
\end{proof}

\paragraph{Attainment: The One-Up-plus-Common-Shock Law (Exact at Finite Parameters, Not
Asymptotic).}
Fix $d > 0$. Let $J$ be uniform on $\{1, \dots, M\}$, let $Y$ be the one-up law of Corollary~1a
($Y_J = (M-1)d$ up, $Y_i = -d$ down for $i \ne J$), and let $W$ be an independent mean-zero
common shock with variance exactly $d^2$. Set
\[
Z_i = Y_i + W.
\]
Then:
\begin{itemize}
\item \textbf{Mean zero:} $\E[Z_i] = 0$.
\item \textbf{Covariances:} for $i \ne j$,
  $\Cov(Y_i, Y_j) = \E[Y_i Y_j] = (2/M) \cdot ((M-1)d)(-d) + ((M-2)/M) \cdot d^2 = -d^2$, so
  $\Cov(Z_i, Z_j) = \Cov(Y_i, Y_j) + \Var W = -d^2 + d^2 = 0$ and the arms are pairwise
  uncorrelated, which fixes the shock variance at $d^2$.
\item \textbf{Variance and maximum:}
  $\Var Z_i = (M-1)d^2 + d^2 = M d^2$, so $V = M^2 d^2$; and $\max_i Z_i = (M-1)d + W$, so
  $\E[\max_i Z_i] = (M-1)d$. The shock adds $M d^2$ to the denominator and nothing to the
  numerator:
  \[
  \E\bigl[\max_i Z_i\bigr] / \sqrt{V} = \frac{(M-1)d}{M d} = \frac{M-1}{M},
  \]
  exactly, at every finite $d$. Equality also holds term-by-term in the master bound:
  $\bar Z = W$, $\E[\bar Z^2] = d^2 = V/M^2$, and
  $\sqrt{(M-1)/M} \cdot \sqrt{V - M\,\E[\bar Z^2]} = \sqrt{(M-1)/M} \cdot \sqrt{M(M-1)} \cdot d
  = (M-1)d$.
\end{itemize}

\paragraph{The $\Cov \ge 0$ Extension.}
Since $\E[\bar Z^2] \ge V/M^2$ holds under pairwise $\Cov \ge 0$ (not only under $\Cov = 0$),
Corollary~1b extends verbatim to the pairwise nonnegatively-correlated class, and is sharp on its
boundary (the uncorrelated subclass). This wider class is also the realistic one: shared
evaluation data induces positive correlation. It is likewise the class in which van Hasselt's
$M = 2$ appendix proof lives: Cantelli's right-hand side is increasing in the variance, so his
substitution of $\Var \hat a_1 + \Var \hat a_2$ for $\Var(\hat a_2 - \hat a_1)$ is valid iff
$\Cov \ge 0$, and his proved $M = 2$ constant $1/2$ is $(M-1)/M$ at $M = 2$.

\begin{remark}[limits of the master bound]
Independence gives the same $\E[\bar Z^2] = V/M^2$ as uncorrelatedness, so the master bound
cannot see past $(M-1)/M$; no variance-decomposition argument can. The independent improvement
(Theorems~2--3) rests on a different mechanism.
\end{remark}

\section{Theorem 2 (Independent Noise, Deterministic Best Arm)}\label{app:thm2}

\begin{thmn}{2}
For independent mean-zero $Z_1, \dots, Z_m$:
\[
\E\bigl[\max(0, Z_1, \dots, Z_m)\bigr] \le \sqrt{m/(2(m+1))} \cdot \sqrt{V},
\quad \text{sharp.}
\]
Equivalently: in the CV problem with $M = m+1$ arms whose best arm has deterministic selection
estimate, the sharp constant is $\sqrt{(M-1)/(2M)}$.
\end{thmn}

\emph{Proof, in five steps.} Write $T = \max(0, Z_1, \dots, Z_m)$.

\subsection{Step 1: Two-Point Reduction per Arm}\label{app:c1}

By Dinkelbach conjugation, maximize $\E[T] - \lambda V$ at the critical $\lambda$. For fixed
other arms, both $\E[T]$ and $V$ are linear in arm $i$'s law, so the per-arm problem maximizes a
linear functional over the constraint set $\{\text{mass } 1,\ \text{mean } 0\}$, whose extreme
points have at most 2 atoms. An arm whose up-atom is non-positive is (by the mean-zero
constraint) the point mass $\delta_0$; it contributes nothing to $T \ge 0$ and is dropped,
reducing $m$; since $m/(2(m+1))$ is increasing in $m$, the bound for the reduced problem implies
the stated one. So without loss of generality arm $i$ takes $u_i > 0$ with probability $x_i$ and
$d_i = -x_i u_i / q_i$ with probability $q_i = 1 - x_i$, with $v_i = x_i u_i^2 / q_i$.

(The fully careful statement of this reduction, covering finiteness of the sup, integrability of
the per-arm integrand, the Bayer--Teichmann representation theorem, and the one-sweep
exact-replacement argument, is written out in Appendix~\ref{app:d1} for Theorem~3; it is the same
Dinkelbach argument, with the deterministic floor entering only through the other-arms maximum
$Z^{\max}_{-i}$.)

\subsection{Step 2: The Exact Formula}\label{app:c2}

Sort the arms so that $u_1 \ge \dots \ge u_m$. Partition by the first up arm: $T = u_i$ iff arm
$i$ is up and all arms $j < i$ are down (arms $j > i$ cannot change the max then, since
$u_j \le u_i$ and their down values are negative); if all arms are down, $T = 0$. Hence
\[
\E[T] = \sum_i x_i u_i s_i, \qquad s_i = \prod_{j<i} q_j.
\]
\subsection{Step 3: The Cauchy--Schwarz Split}\label{app:c3}

Factor each term and apply Cauchy--Schwarz:
\begin{align*}
\E[T] &= \sum_i \Bigl[ u_i \sqrt{x_i/q_i} \Bigr] \cdot \Bigl[ \sqrt{x_i q_i} \cdot s_i \Bigr]\\
&\le \sqrt{\textstyle\sum_i x_i u_i^2/q_i} \cdot \sqrt{\textstyle\sum_i x_i q_i s_i^2}\\
&= \sqrt{V} \cdot \sqrt{f(x)}, \qquad f(x) = \sum_i x_i q_i s_i^2,
\end{align*}
with equality at $u_i \propto q_i s_i$. This choice is \textbf{sort-consistent}: since
$s_{i+1} = s_i q_i$,
\[
u_{i+1}/u_i = (q_{i+1} s_{i+1}) / (q_i s_i) = q_{i+1} \le 1,
\]
so the maximizing $u$ respects the assumed order $u_1 \ge \dots \ge u_m$, and the split loses
nothing.

\subsection{Step 4: The Recursion and Its Exact Solution}\label{app:c4}

Since $s_1 = 1$ and $s_i = q_1 \cdot \prod_{1<j<i} q_j$ for $i \ge 2$,
\[
f(x) = x_1 q_1 + q_1^2 \cdot f(x_2, \dots, x_m),
\]
so $\max_x f = G_m$ where
\[
G_m = \max_{x \in [0,1]} \bigl[\, x(1-x) + (1-x)^2 G_{m-1} \,\bigr], \qquad G_0 = 0.
\]

\textbf{Claim:} $G_m = m/(2(m+1))$, with optimizer $x^* = 1/(m+1)$.

\begin{proof}[Proof by induction]
Base: $G_0 = 0 = 0/2$. Assume $G_{m-1} = (m-1)/(2m)$. The objective
$x(1-x) + (1-x)^2 (m-1)/(2m)$ is a concave quadratic in $x$; setting its derivative to zero,
\begin{align*}
&(1 - 2x) - \tfrac{m-1}{m}(1-x) = 0\\
&\quad\iff m(1-2x) = (m-1)(1-x)
\iff x(m+1) = 1,
\end{align*}
so $x^* = 1/(m+1) \in (0,1)$. The value there is
\begin{align*}
x^*(1-x^*) + (1-x^*)^2 G_{m-1}
&= \tfrac{m}{(m+1)^2} + \bigl(\tfrac{m}{m+1}\bigr)^2 \tfrac{m-1}{2m}\\
&= \frac{2m + m(m-1)}{2(m+1)^2}\\
&= \frac{m(m+1)}{2(m+1)^2} = \frac{m}{2(m+1)}. \qedhere
\end{align*}
\end{proof}

Combining Steps 3--4: $\E[T] \le \sqrt{V} \cdot \sqrt{G_m} = \sqrt{m/(2(m+1))} \cdot \sqrt{V}$.

\subsection{Step 5: Attainment (The Cascade Law)}\label{app:c5}

Take arm $k = 1, \dots, m$:
\[
\begin{gathered}
u_k = \tfrac km \ \text{with probability}\ x_k = \tfrac1{k+1},\\
\text{hence}\ d_k = -\tfrac{x_k u_k}{q_k} = -\tfrac1m\ \text{for every } k.
\end{gathered}
\]
The supremum is attained exactly, for every $m$, by a two-line computation. Each arm has
\[
\Var Z_k = \frac{x_k u_k^2}{q_k}
= \frac{1}{k+1} \cdot \frac{k^2}{m^2} \cdot \frac{k+1}{k} = \frac{k}{m^2},
\]
so $V = \sum_{k=1}^m k/m^2 = (m+1)/(2m)$. The cascade is indexed by increasing up-value
($u_k = k/m$), so arm $k$ occupies sorted position $m+1-k$, and the head survival factor of
Appendix~\ref{app:setup} at that position is the product of the $q_j$ over the arms with larger
up-value, that is over $j > k$. With $q_j = 1 - 1/(j+1) = j/(j+1)$ the product telescopes:
\[
s_{(k)} = \prod_{j>k} q_j = \prod_{j=k+1}^{m} \frac{j}{j+1} = \frac{k+1}{m+1},
\]
so each term of the exact formula is
\[
x_k u_k s_{(k)} = \frac{1}{k+1} \cdot \frac{k}{m} \cdot \frac{k+1}{m+1} = \frac{k}{m(m+1)},
\]
and
\[
\E[T] = \sum_{k=1}^m \frac{k}{m(m+1)} = \frac{1}{2} \quad \text{identically in } m.
\]
Hence $\E[T]^2/V = (1/4) \cdot 2m/(m+1) = m/(2(m+1)) = G_m$ exactly. The same configuration
appears as the Theorem-3 attaining law with the floor realized by a deterministic-zero arm
(Appendix~\ref{app:d5}). \qed

\section{Theorem 3 (Independent Noise, General)}\label{app:thm3}

\begin{thmn}{3}
For independent mean-zero $Z_1, \dots, Z_M$:
\[
\sup \E\bigl[\max_i Z_i\bigr] / \sqrt{V} \;=\; \sqrt{(M-1)/(2M)} \;=\; \sqrt{G_{M-1}},
\]
i.e.\ the Theorem-2 value with one fewer arm: the optimal role of one arm is as a deterministic
floor. The sup is attained exactly (Appendix~\ref{app:d5}).
\end{thmn}

The proof is a chain of four reductions plus the attainment lemma:
\begin{align*}
&\text{two-point reduction (\ref{app:d1})} \to \text{convexity (\ref{app:d2})} \to {}\\
&\text{head recursion (\ref{app:d3})} \to \Phi_t \le G_t \ \text{(\ref{app:d4})};
\quad \text{attainment (\ref{app:d5})}.
\end{align*}
Throughout, only upper bounds ($\le$) are needed in \ref{app:d1}--\ref{app:d4}; equality follows
from \ref{app:d5}.

\subsection{The Two-Point Reduction}\label{app:d1}

\begin{uproposition}[two-point reduction]
\[
R^* := \sup\{ \E[\max_i Z_i] : Z_i \text{ independent, mean-zero, } \textstyle\sum \Var Z_i \le 1 \}
\]
is unchanged if every arm is restricted to laws with at most two atoms.
\end{uproposition}

\begin{proof}
\textbf{Step 0 (normalization and finiteness).}
$\E[\max_i Z_i] \ge \max_i \E[Z_i] = 0$, and scaling $Z \to cZ$ scales the objective by $c$ and
the variance by $c^2$, so the sup over $\sum \Var \le 1$ equals (as suprema) the sup at
$\sum \Var = 1$, and the ratio form $\E[\max]/\sqrt{\sum \Var}$ has the same value. $R^*$ is
finite before anything else is argued: Corollary~1a (Samuelson pointwise plus Cauchy--Schwarz;
no independence used) gives $\E[\max_i Z_i] \le \sqrt{(M-1)/M} \cdot \sqrt{V}$ for every
mean-zero vector, so
\[
0 \le R^* \le \sqrt{(M-1)/M} < 1.
\]
Assume $M \ge 2$ throughout Steps 1--5 (at $M = 1$ the theorem is the trivial $0 = 0$ and
$Z^{\max}_{-i}$ below would be a maximum over an empty set); for $M \ge 2$, $R^* > 0$ (the arm $\pm 1$
with probability $1/2$ each, plus a zero arm, gives $1/2$).

\textbf{Step 1 (Dinkelbach conjugation).}
For $\lambda > 0$ let
\[
\Phi(\lambda) = \sup\Bigl\{ \E\bigl[\max_i Z_i\bigr] - \lambda \cdot \sum \Var Z_i \Bigr\}
\]
over independent mean-zero laws, unconstrained. For any $Z$ with $\sum \Var = s$:
$\E[\max] \le R^* \sqrt{s}$ (finite by Step 0), so
\[
\E[\max] - \lambda s \le \max_{s \ge 0} \bigl( R^* \sqrt{s} - \lambda s \bigr)
= R^{*2}/(4\lambda);
\]
conversely, scaling a near-optimal $Z$ to $s = (R^*/2\lambda)^2$ shows equality. So
$\Phi(\lambda) = R^{*2}/(4\lambda) < \infty$, and determining $\Phi$ on the two-point class
determines $R^*$ on it, unambiguously, via $R^* = 2\sqrt{\lambda \Phi(\lambda)}$, which needs
only $R^* \ge 0$.

\textbf{Step 2 (per-arm linearity).}
Fix $\lambda$ and all arms except $i$, and let $Z^{\max}_{-i} = \max_{j \ne i} Z_j$ (independent of
$Z_i$). Define
\[
g(z) = \E\bigl[ \max(z, Z^{\max}_{-i}) \bigr].
\]
$g$ is a 1-Lipschitz convex function with $g(z) \to \E[Z^{\max}_{-i}]$ as $z \to -\infty$ and
$g(z) \sim z$ as $z \to +\infty$ (indeed $g(z) - z = \E[(Z^{\max}_{-i} - z)^+] \downarrow 0$). The
Fubini step behind $g$ is licensed: $|\max(z, Z^{\max}_{-i})| \le |z| + |Z^{\max}_{-i}|$, both integrable
(finite variance plus mean zero implies $L^1$), product measure, so Tonelli/Fubini applies. The
per-arm objective is then
\[
J_i(\mu) = \int \psi \, d\mu + \text{const}, \qquad \psi := g - \lambda z^2
\]
(defined as one integral, not a difference), which is linear in $\mu$; the constraint set is
$\{ \mu \in P(\mathbb{R}) : \int z \, d\mu = 0 \}$.

\textbf{Step 3 (integrability; no compactness needed).}
$\psi$ is bounded above: on the left, $g \to \E[Z^{\max}_{-i}]$, so $\psi \le C - \lambda z^2$; on the
right, $g(z) - z = \E[(Z^{\max}_{-i} - z)^+] \downarrow 0$, so $\psi \le C + z - \lambda z^2$. Hence
$\sup J_i < \infty$, $J_i$ is well-defined in $[-\infty, \infty)$ for every mean-zero $\mu$
(measures without a second moment score $-\infty$), and any $\mu$ with $J_i(\mu) > -\infty$ has
$\psi \in L^1(\mu)$. That is all that Step 4's general-integrand theorem requires: no
compact-support detour is necessary.

The compact-support route via Tchakaloff's theorem is therefore available but not needed, and we do not take it.

\textbf{Step 4 (Bayer--Teichmann plus linear-program basic solutions).}
Consider the moment map
\[
\mu \mapsto \Bigl( \int d\mu, \ \int z \, d\mu, \ \int \psi \, d\mu \Bigr).
\]
Apply Corollary~2 of \citet{bayerteichmann2006} with $\Omega = \mathbb{R}$, $A$ any full-measure
set, and $\varphi = (1, z, \psi) : \mathbb{R} \to \mathbb{R}^3$: for a positive measure $\mu$
concentrated in $A$ and a Borel map $\varphi : \Omega \to \mathbb{R}^N$ whose push-forward
$\varphi_* \mu$ has first moments, there exist an integer $1 \le k \le N$, points
$\omega_1, \dots, \omega_k \in A$ and weights $\lambda_i > 0$ with
$\int_\Omega \varphi_j \, d\mu = \sum_{i=1}^k \lambda_i \varphi_j(\omega_i)$ for every $j$, under
integrability alone. The hypotheses hold here: $1, z, \psi$ are Borel
and $\mu$-integrable; $1$ because $\mu$ is a probability measure; $z \in L^1(\mu)$ because the
mean-zero constraint presupposes it; $\psi \in L^1(\mu)$ may be assumed by Step 3 ($\psi$ is
continuous and bounded above). Conclusion: the value vector is achieved by a measure with at most
$N = 3$ atoms, of mass exactly $1$ (the $\varphi_1 = 1$ coordinate) and mean exactly $0$ (the
$\varphi_2 = z$ coordinate).

The argument does not depend on the exact atom count from the citation: among measures on any
finite atom set, maximizing the linear objective $\int \psi \, d\mu$ under the two equality
constraints ($\text{mass} = 1$, $\text{mean} = 0$) is a linear program (LP) whose constraint
matrix $[[1, 1, \dots], [z_1, z_2, \dots]]$ has rank $\le 2$ and whose feasible set is a compact
subset of the simplex; hence the LP is never unbounded, and an optimal vertex with at most 2
active atoms exists (rank-1 degeneracy forces a one-point measure, which, being mean-zero, is
$\delta_0$). Combining: the per-arm sup is approached within two-point mean-zero laws. (The
underlying convexity fact is that the barycenter of an integrable random vector lies in the
convex hull, not just the closed hull, of its support; this is the probability-measure
consequence that \citet{bayerteichmann2006} derive from their Theorem~1.)

\textbf{Step 5 (one sweep, exact replacements).}
The joint objective is $J_i(\mu_i) + c_i$, where both $g$ and
$c_i = -\lambda \sum_{j \ne i} \Var Z_j$ depend only on the other arms. Replace arms one at a
time by an \textbf{exact} LP optimum: by Steps 3--4, for each arm there is a two-point mean-zero
law whose $J_i$ value is $\ge J_i(\mu_i)$ (the finitely-atomic representer followed by the LP
vertex; both exact, no per-arm $\varepsilon$). Each replacement does not decrease the joint
value, and a previously replaced arm's \emph{law} stays supported on at most two atoms (possibly
the one-point $\delta_0$; later replacements change only its role in $Z^{\max}_{-j}$, and support size
is a property of the law itself). A later replacement may destroy the per-arm optimality of an
earlier one; this is immaterial, because the argument uses only two facts about each
replacement: it does not decrease the joint value, and it leaves the replaced arm two-point. One
sweep of $M$ exact replacements from an
$\varepsilon$-optimal starting configuration gives
$\Phi_{\mathrm{2pt}}(\lambda) \ge \Phi(\lambda) - \varepsilon$; with the trivial $\le$,
$\Phi_{\mathrm{2pt}} = \Phi$, and Step 1 run inside the two-point class (which is scaling-closed)
gives $R_{\mathrm{2pt}} = R^*$. The sweep terminates in $M$ steps.
\end{proof}

\begin{remark}[degenerate arms]
A one-point mean-zero arm is $\delta_0$; if any arm is degenerate, the objective is
$\E[\max(0, Z_{\text{rest}})]$ with $m \le M - 1$ random arms and the same $V$, which Theorem~2
bounds by $\sqrt{G_m} \cdot \sqrt{V} \le \sqrt{G_{M-1}} \cdot \sqrt{V}$ ($G_m = m/(2(m+1))$ is
increasing in $m$). So it suffices to bound configurations in which every arm is
two-point, which is what \ref{app:d2}--\ref{app:d4} do.
\end{remark}

\subsection{The Convexity Step}\label{app:d2}

Work in the two-point class with $\sum \Var = 1$, arms sorted by up value, and the notation of
Appendix~\ref{app:setup} ($w_i = x_i u_i$, $s_i$, $Q$, $P_k$). The formula for $E(w)$ below is
derived under the sort order $u_1 \ge \dots \ge u_M$; the maximization over the variance
ellipsoid afterwards drops the order constraint, which can only enlarge the supremum, and an
upper bound is all that is required. Partition by the first up arm
as in Appendix~\ref{app:c2}, except that the all-down event (probability $Q$) now
contributes the largest \emph{down} value $\max_i d_i = -\min_i (w_i/q_i)$ instead of $0$; the
objective is
\[
E(w) = \sum_i w_i s_i - Q \cdot \min_i (w_i / q_i).
\]
Since $-\min = \max$ of the negatives, $E(w) = \max_k L_k(w)$ with
\[
L_k(w) = \sum_i w_i s_i - Q \cdot w_k / q_k,
\]
a maximum of $M$ \textbf{linear} functionals of $w$, hence convex; and $E$ is homogeneous of
degree 1 in $w$. For fixed $x$, the variance constraint
$\sum_i v_i = \sum_i w_i^2/(x_i q_i) \le 1$ is an ellipsoid in $w$, so
\[
\sup_{\text{ellipsoid}} E = \max_k \sup_{\text{ellipsoid}} L_k = \max_k \sqrt{h_k(x)},
\]
with
\[
h_k(x) := \sum_{i \ne k} x_i q_i s_i^2 + x_k q_k s_k^2 (1 - P_k)^2,
\]
where $P_k = \prod_{j>k} q_j$ (the coefficient of $w_k$ in $L_k$ is
$s_k - Q/q_k = s_k(1 - P_k)$, using $Q = s_k q_k P_k$; equivalently, the expansion uses
$1 - P(2-P) = (1-P)^2$). Because the supremum is taken as a maximum over the $k$ attaining
arms rather than through a first-order condition, ties among them, subgradients and boundary cases
play no role.

If $k$ is last in sort order, $P_k = 1$ and $h_k = f_{M-1}(x) \le G_{M-1}$ immediately
(Theorem~2, Step~\ref{app:c4}). The remaining task is $\sup_x h_k(x) \le G_{M-1}$ for every $k$.

\subsection{The Head Recursion}\label{app:d3}

Decompose $h_k$: with $A = \sum_{i<k} x_i q_i s_i^2$ the head sum, $S = s_k$, tail value
$\Theta = f_t(y)$ over the $t = M - k$ arms after $k$ (in their own coordinates $y$, with survivals
computed inside the tail), and tail survival $P = P_k$,
\[
h_k = A + S^2 \cdot \bigl[\, x_k q_k (1 - P)^2 + q_k^2 \Theta \,\bigr]
\]
(for $i > k$, $s_i = s_k q_k \cdot \prod_{k<j<i} q_j$, which yields the factor $S^2 q_k^2$ on the
tail sum). The bracket equals
\[
\Phi_t(x_k, y) := x_k(1 - x_k)(1 - P(y))^2 + (1 - x_k)^2 f_t(y),
\]
the key quantity of Appendix~\ref{app:d4}. Maximizing the head over its $k-1$ arms is the
recursion $R(c) = \max_x \bigl[ x(1-x) + (1-x)^2 c \bigr]$ applied $k-1$ times (peel the first
arm as in Appendix~\ref{app:c4}: $A + S^2 c = x_1 q_1 + q_1^2 [A' + S'^2 c]$), $R$ is increasing
in $c$ (the coefficient $(1-x)^2$ is nonnegative), and $R(G_j) = G_{j+1}$, which is the
Theorem-2 recursion. Hence
\[
\sup_x h_k \le R^{k-1}\bigl( \sup \Phi_{M-k} \bigr),
\]
and \textbf{only $\le$ is needed}; equality holds numerically but is not claimed. Given
$\Phi_t \le G_t$ (Appendix~\ref{app:d4}),
\[
\sup_x h_k \le R^{k-1}(G_{M-k}) = G_{M-1} \qquad \text{for every } k = 1..M.
\]
(The composition identity holds because $R(G_j) = G_{j+1}$ for every $j$: this is the $m = j+1$
instance of the recursion solved in Appendix~\ref{app:c4}, so $R^{k-1}(G_{M-k}) = G_{M-1}$
follows by induction on $k$.) Combining
\ref{app:d1}--\ref{app:d3} with \ref{app:d4}:
\[
\begin{aligned}
\sup \E\bigl[\max_i Z_i\bigr]/\sqrt{V} &\le \max_k \sqrt{\sup\nolimits_x h_k}\\
&\le \sqrt{G_{M-1}} = \sqrt{(M-1)/(2M)}.
\end{aligned}
\]

\subsection{The Key Inequality $\Phi_t \le G_t$}\label{app:d4}

\begin{uproposition}
For a $t$-arm tail ($t \ge 1$),
\[
\Phi_t(x, y) = x(1-x)(1 - P(y))^2 + (1-x)^2 f_t(y) \le G_t,
\]
where $P(y) = \prod_j (1 - y_j)$ and $f_t(y) = \sum_j y_j (1 - y_j) (s'_j)^2$ with $s'_j$ the
within-tail survivals.
\end{uproposition}

\emph{Proof, in four steps.}

\textbf{Step 1 (the tail trade-off curve has a closed form).}
Define
\[
\Psi_t(P) = \sup\bigl\{ f_t(y) : \textstyle\prod_j (1 - y_j) = P \bigr\},
\]
the best tail value at fixed tail survival. Then
\[
\Psi_t(P) = \frac{(1 - P) \cdot \bigl[ (t-1) + (t+1)P \bigr]}{2t}.
\]

\begin{proof}[Proof by induction]
Base $t = 1$: $\Psi_1(P) = P(1-P)$ directly (one arm, $y_1 = 1 - P$), matching the formula. For
the step, peel the first tail arm ($r = 1 - y_1$, so the remaining constraint is
$\prod_{j \ge 2} (1 - y_j) = P/r$):
\[
\Psi_t(P) = \max_{r \in [P, 1]} \bigl[\, (1-r)r + r^2 \Psi_{t-1}(P/r) \,\bigr].
\]
Substituting the inductive form, the objective is a downward parabola in $r$ (leading coefficient
$-t/(2(t-1)) < 0$) with unique critical point
\[
r^* = \frac{t - 1 + P}{t},
\]
which is feasible: $r^* \ge P \iff (t-1)(1-P) \ge 0$, and $r^* \le 1 \iff P \le 1$. Explicitly,
substituting $\Psi_{t-1}(P/r) = (r-P)((t-2)r + tP)/(2(t-1)r^2)$ gives the objective
\[
r - r^2 + \frac{(r-P)\bigl((t-2)r + tP\bigr)}{2(t-1)},
\]
a quadratic in $r$ with coefficients
\[
a = \frac{-t}{2(t-1)}, \qquad
b = \frac{t-1+P}{t-1}, \qquad
c = \frac{-tP^2}{2(t-1)}.
\]
Its value at the vertex $r^* = -b/(2a) = (t-1+P)/t$ is $c - b^2/(4a)$, and the numerator
factors:
\[
\begin{aligned}
(t-1+P)^2 - t^2 P^2 &= (t-1+P-tP)(t-1+P+tP)\\
&= (t-1)(1-P)\\
&\qquad \cdot \bigl[ (t-1) + (t+1)P \bigr],
\end{aligned}
\]
so the vertex value is $(1-P)\bigl[(t-1)+(t+1)P\bigr]/(2t)$, which is the claimed form.

\emph{The $P = 0$ branch.} The peeling identity presumes $r > 0$. At $P = 0$ the excluded
$r = 0$ branch ($y_1 = 1$) gives $f_t = y_1(1 - y_1) + 0 = 0$, which never attains the max for
$t \ge 2$. Moreover $\Psi_t(0) = G_{t-1}$ is attained as a value, not merely a supremum: $f_t$
is a polynomial, well-defined at $y_j = 1$, and placing an always-firing arm last reduces $f_t$
to $f_{t-1}$ of the remaining arms, whose maximum is $G_{t-1}$ by Appendix~\ref{app:c4}.

Two checks of the closed form: at $P = 1/(t+1)$, $1 - P = t/(t+1)$ and
$(t-1) + (t+1)P = t$, so $\Psi_t = t/(2(t+1)) = G_t$, matching the telescoped survival of the
optimal schedule; at $P = 0$, $\Psi_t = (t-1)/(2t) = G_{t-1}$, consistent with an always-firing
arm placed last costing nothing.
\end{proof}

\textbf{Step 1$'$ (order-independence; why the upper envelope is safe).}
$f_t$ is order-dependent, but the $u$-variables were already eliminated by the Cauchy--Schwarz
step, so no sort constraint binds on $y$; and $\Psi_t$ takes the sup over \emph{all} ordered
tuples, which dominates the specific ordered configuration appearing in $h_k$. The bound
$f_t(y) \le \Psi_t(P(y))$ therefore holds pointwise, and only this pointwise domination is used
below.

\textbf{Step 2 (vertex analysis).}
For fixed tail $y$, Step 1$'$ gives
\[
\begin{gathered}
\Phi_t(x, y) \le F(x) := \alpha \, x(1-x) + \beta \, (1-x)^2,\\
\alpha = (1-P)^2, \qquad \beta = \Psi_t(P).
\end{gathered}
\]
$F$ is a quadratic in $x$ with $F(0) = \beta$, $F(1) = 0$, and vertex
$x^* = (\alpha - 2\beta)/(2(\alpha - \beta))$ when $\beta < \alpha$. Cases:
\begin{itemize}
\item \textbf{$\beta \ge \alpha$:} the parabola opens upward (coefficient
  $\beta - \alpha \ge 0$), so the max over $[0,1]$ is at an endpoint;
  $F(0) = \beta = \Psi_t(P) \le G_t$ (Step 1's form is maximized at $P = 1/(t+1)$ with value
  $G_t$) and $F(1) = 0$.
\item \textbf{$\alpha/2 \le \beta < \alpha$:} downward parabola with vertex
  $x^* = (\alpha - 2\beta)/(2(\alpha - \beta)) \le 0$, so the max on $[0,1]$ is at $x = 0$,
  $F = \beta \le G_t$.
\item \textbf{$\beta < \alpha/2$ (interior vertex):} the vertex value is
  $F^* = \beta + (\alpha - 2\beta)^2/(4(\alpha - \beta))$, and
  \[
  F^* \le G_t \iff \beta \le \alpha - \alpha^2/(4 G_t).
  \]
  \begin{itemize}
  \item Sub-case $\alpha \le 2 G_t$:
    $\alpha - \alpha^2/(4 G_t) \ge \alpha - \alpha/2 = \alpha/2 > \beta$, so $F^* \le G_t$.
  \item Sub-case $\alpha > 2 G_t$, i.e.\ $a := 1 - P > \sqrt{t/(t+1)}$: the one nontrivial
    branch, handled in Step 3.
  \end{itemize}
\end{itemize}

\textbf{Step 3 (the nontrivial branch is a factored cubic).}
With $\Psi_t(1-a) = a(2t - (t+1)a)/(2t)$ and $1/(4 G_t) = (t+1)/(2t)$, the required inequality
$\beta \le \alpha - \alpha^2/(4 G_t)$ reads $\Psi_t(1-a) \le a^2 - (t+1)a^4/(2t)$, and
\[
\begin{aligned}
&(2t/a) \cdot \bigl[\, a^2 - (t+1)a^4/(2t) - \Psi_t(1-a) \,\bigr]\\
&\quad = -(a - 1) \cdot \bigl( (t+1)a^2 + (t+1)a - 2t \bigr)
\end{aligned}
\]
(an identity verified by expanding both sides). On $a \in [\sqrt{t/(t+1)}, 1]$: the factor
$(a - 1) \le 0$; the second factor at the left endpoint is
$(t+1) \cdot t/(t+1) + (t+1)\sqrt{t/(t+1)} - 2t = \sqrt{t(t+1)} - t > 0$ and is increasing in
$a$, so the product is $\ge 0$ throughout, i.e.\ $\Psi_t(1-a) \le a^2 - a^4/(4 G_t)$
where it is needed, with equality only at $a = 1$. \qed

\textbf{Step 4 (endpoints meet, and the equality branch).}
At $a = 1$ ($P = 0$):
\[
\Psi_t(0) = G_{t-1} = 1 - \frac{1}{4 G_t}
\]
(both sides equal $(t-1)/(2t)$): the constraint curve and the trade-off curve touch at
the always-firing-tail configuration. The equality branch at $a = 1$ for general $t$ is
\[
x^* = 1/(t+1) \quad \text{with an always-firing \textbf{last} tail arm.}
\]
This is the regime in which the tail always fires and arm $k$ itself takes the role of the
deterministic floor. \qed

This completes the chain: \ref{app:d1} (two-point reduction) $\to$ \ref{app:d2}
($\max_k \sqrt{h_k}$) $\to$ \ref{app:d3} ($R^{k-1}$ composition, $\le$ only) $\to$ \ref{app:d4}
($\Phi_t \le G_t$) gives $\sup \E[\max_i Z_i]/\sqrt{V} \le \sqrt{(M-1)/(2M)}$ for independent
mean-zero noise. Equality is Appendix~\ref{app:d5}.

\subsection{The Attainment Lemma}\label{app:d5}

\begin{attainmentlemma}
The Theorem-3 sup is attained exactly for every $M$: the $(M-1)$-arm cascade of Theorem~2 (arm
$k$: $u_k = k/(M-1)$ with probability $1/(k+1)$, else $-1/(M-1)$) plus one
\textbf{deterministic-zero arm} is a legitimate independent $M$-arm mean-zero configuration, and
$\E[\max_i Z_i] = \E[\max(0, Z_1, \dots, Z_{M-1})]$ is the Theorem-2 quantity with
$m = M - 1$, so by the computation in Appendix~\ref{app:c5} (with $m = M - 1$: $\E[T] = 1/2$,
$V = M/(2(M-1))$),
\[
\E[\max]^2 / V = \frac{1}{4} \cdot \frac{2(M-1)}{M} = \frac{M-1}{2M}.
\]
The bias-transfer form $\E[g_I]$ needs a limiting construction, given below; the ratio
$\E[\max_i Z_i]/\sqrt{V}$ itself does not.
\end{attainmentlemma}

\paragraph{The Limiting Constructions for $\E[g_I]$.}
The gaps must be placed just \emph{below} the up-values of the extremal law, not near zero. Give
the deterministic arm the top mean, $g_0 = 0$, and arm $k$ the gap
$g_k = u_k - \varepsilon u_k$ with $\varepsilon \in (0,1)$, where $u_k = k/(M-1)$ is its up-value.
Arm $k$ then beats the floor exactly when it is up, since $Z_k - g_k = \varepsilon u_k > 0$ up and
$-1/(M-1) - g_k < 0$ down; and among the up arms the largest $\varepsilon u_k$ wins, which is the
same arm that attains $\max(0, Z_1, \dots, Z_{M-1})$. Hence
\[
\E[g_I] \;=\; (1 - \varepsilon)\, \E\bigl[\max(0, Z_1, \dots, Z_{M-1})\bigr]
\;\xrightarrow[\varepsilon \to 0^+]{}\; \E\bigl[\max_i Z_i\bigr],
\]
with $V$ unchanged, since the means do not enter it. The ratio $\E[g_I]/\sqrt{V}$ therefore rises
to the constant as $\varepsilon \to 0^+$, that is as every alternative's deficit approaches the
gain its own noise can supply and selection becomes a near-tie. The margins $\varepsilon u_k$ must
keep the order of the $u_k$: with gaps just below the $u_k$ but margins in another order the limit
is smaller, for instance $5/12 < 1/2$ at $M = 3$.

At $M = 2$ this is the whole picture in closed form. With the floor $Z_1 = 0$, $Z_2 = \pm s$ and
gap $g$ on the second arm, the second arm is selected exactly when it is up and $g < s$, so
\[
\frac{\E[g_I]}{\sqrt{V}} = \frac{g}{2s} \quad (g < s), \qquad
\frac{g}{4s} \quad (g = s), \qquad 0 \quad (g > s).
\]
The supremum $1/2$ is approached as $g \to s^-$. At $g \to 0$ the ratio goes to zero: a pool of
equally good alternatives has no bias to transfer, because every selection is then correct.
Vanishing gaps are the wrong limit.

\paragraph{The One-Up Laws.}
The same mechanism carries the constants of Corollaries~1a and~1b over to the $\E[g_I]$ form, but
the gaps must rise to $Md$, the lead of an up arm over the optimal arm (which is then down), and
not to the up-value $(M-1)d$. Give arm~$1$ the top mean and every other arm the gap $g$. When
$J \ne 1$ the up arm $J$ beats every other non-optimal arm and beats arm~$1$ exactly when
$g < Md$; when $J = 1$ arm~$1$ wins; and the common shock of Corollary~1b shifts every score
equally. With ties broken uniformly, for both laws,
\[
\E[g_I] = \tfrac{M-1}{M}\, g \quad (g < Md), \qquad \tfrac{M-1}{2M}\, g \quad (g = Md), \qquad
0 \quad (g > Md),
\]
with $V$ unchanged. Hence $\E[g_I] \to (M-1)d = \E[\max_i Z_i]$ as $g \to Md^-$, and
$\E[g_I]/\sqrt{V}$ rises to $\sqrt{(M-1)/M}$ and to $(M-1)/M$ respectively; gaps at the up-value
$(M-1)d$ give only the fraction $(M-1)/M$ of each constant.

\paragraph{Necessity of a Deterministic Arm at $M = 2$ (Proved).}
$\E[\max(Z_1, Z_2)] = \E|Z_1 - Z_2|/2 \le \sqrt{v_1 + v_2}/2$, with equality iff $|Z_1 - Z_2|$
is a.s.\ constant; for independent arms this forces one arm deterministic ($Z_1 = 0$ a.s.,
$Z_2 = \pm s$): the deterministic-floor structure appears by itself.

\begin{remark}[conjecture]
Whether every attaining configuration at general $M$ must contain a deterministic arm is stated
as a conjecture; Sec.~\ref{sec:limitations} records its status.
\end{remark}

\section{Formal Verification in Lean 4}\label{app:lean}

The formalization, which will be released with the code, covers every theorem, corollary and
lemma of the paper in Lean~4 with Mathlib (toolchain and Mathlib \texttt{v4.34.0-rc1}; 11 files,
about 7{,}900 lines). It contains no \texttt{sorry}, and every theorem depends only on Lean's
standard axioms \texttt{propext}, \texttt{Classical.choice} and \texttt{Quot.sound}. It builds
with \texttt{lake build}.

\paragraph{What Is Formalized.}
Random-variable statements are made on an arbitrary probability space $(\Omega, P)$ for
$Z : \mathrm{Fin}\,M \to \Omega \to \mathbb{R}$, with hypotheses \texttt{MemLp (Z i) 2 P} (finite
variance), $\int Z_i \, dP = 0$ and, where the class requires it, \texttt{iIndepFun Z P}
(mutual independence). The maximum is the pointwise supremum $\omega \mapsto \sup_i Z_i(\omega)$
and $V$ is the sum of the Mathlib variances. Sharpness is formalized as a greatest element: for
Theorems~2 and~3 the statement \texttt{IsGreatest} asserts both that the constant bounds the ratio
$\E[\max_i Z_i]/\sqrt{V}$ over all admissible laws and that some law attains it.

\paragraph{Correspondence.}
Table~\ref{tab:lean} lists the main Lean names; the \texttt{README} of the released formalization
gives the full list, including the paper-specific algebra of Appendices~\ref{app:thm2} and~\ref{app:thm3}
and the closed forms behind Tables~\ref{tab:graded} and~\ref{tab:gaussian}.

\begin{table}[h]
\centering
\caption{Paper results and the Lean theorems that verify them.}\label{tab:lean}
\footnotesize
\setlength{\tabcolsep}{3pt}
\begin{tabular}{lll}
\toprule
paper & Lean & file \\
\midrule
Lemma~1, bias ranges & \texttt{Transfer.transfer}, \texttt{me\_bias\_mem}, \texttt{cv\_bias\_mem} & \texttt{Transfer} \\
Theorem~1 & \texttt{Master.master\_bound} & \texttt{Master} \\
Corollaries~1a, 1b (bounds) & \texttt{cor1a\_bound}, \texttt{cor1a\_eq\_iff}, \texttt{cor1b\_bound}, \texttt{cor1b\_strict} & \texttt{Master} \\
Corollaries~1a, 1b (attaining laws) & \texttt{Attain1.Y\_ratio}, \texttt{pair\_ratio}, \texttt{shock\_attains} & \texttt{Attain1} \\
Theorem~2 & \texttt{theorem2}, \texttt{theorem2\_measure}, \texttt{theorem2\_sup} & \texttt{Independent} \\
Theorem~3 & \texttt{theorem3}, \texttt{theorem3\_measure}, \texttt{theorem3\_sup} & \texttt{Independent} \\
Theorem~3 at $M = 2$, deterministic arm & \texttt{thm3\_M2\_bound}, \texttt{thm3\_M2\_equality} & \texttt{Necessity} \\
Two-point reduction (App.~\ref{app:d1}) & \texttt{reduction\_ratio\_rv}, \texttt{reduction} & \texttt{Reduction} \\
Cascade attainment (App.~\ref{app:c5}, \ref{app:d5}) & \texttt{TwoPoint.cascade\_*}, \texttt{cascadeZ\_*} & \texttt{TwoPoint} \\
Corollary~2 & \texttt{Transfer.cor2}, \texttt{cor2\_strict\_iff}, \texttt{cor2\_me} & \texttt{Transfer} \\
$\E[g_I]$ limits (App.~\ref{app:d5}) & \texttt{cascade\_limit}, \texttt{m2\_closed\_form}, \texttt{ouEgap\_*} & \texttt{Transfer} \\
\bottomrule
\end{tabular}
\end{table}

\paragraph{Where the Formal Proofs Take Shorter Routes.}
Two steps of the independent case are formalized along routes different from, and shorter than,
the ones written in this appendix. The two-point reduction of Appendix~\ref{app:d1} is proved
without \citet{bayerteichmann2006}: an elementary supporting-line lemma (a function bounded by $B$
on every mean-zero two-point law satisfies $\varphi(z) \le B + cz$ for some $c$) combined with
Fubini's theorem arm by arm. The two-point bound of Theorem~3 is proved by one induction that
removes the arm with the largest up-value and applies a two-term Cauchy--Schwarz step, and the
inequality $\Phi_t \le G_t$ of Appendix~\ref{app:d4} follows, once $f_t$ is replaced by its
closed-form bound $\Psi_t$, from the sum-of-squares certificate
\[
\begin{aligned}
&2t(t+1)\bigl(G_t - x(1-x)(1-P)^2 - (1-x)^2\Psi_t(P)\bigr)\\
&\qquad = \bigl(t - (t+1)(1-x)(1-P)\bigr)^2 + 2t(t+1)\, xP(1-x)(1-P),
\end{aligned}
\]
valid for $t \ge 1$ and nonnegative on $x, P \in [0, 1]$.
The route written here (ellipsoid, maximum of $M$ values, vertex analysis and factored cubic) is
formalized separately as well, so both proofs are checked.

\end{document}